\documentclass[journal]{IEEEtran} 

\usepackage[cmex10]{amsmath}
\usepackage{amssymb}
\usepackage{color}
\usepackage{wasysym} 

\newcommand{\bitem}{\begin{itemize}}
\newcommand{\eitem}{\end{itemize}}

\newcommand{\bpm}{\begin{pmatrix}}   
\newcommand{\epm}{\end{pmatrix}}

\newcommand{\bq}{\begin{equation}}
\newcommand{\eq}{\end{equation}}

\usepackage{booktabs}
\usepackage{graphicx}
\usepackage{bm}
\usepackage{algorithmic}
\usepackage{color}

\usepackage[tight,footnotesize]{subfigure}

\usepackage{epsfig}
\usepackage{epstopdf}

\usepackage[draft]{todonotes}   

\begin{document}
%
 
\title{Learning under Distributed Weak Supervision}

%
\author{Martin Rajchl, Matthew~C.~H.~Lee, Franklin~Schrans, Alice~Davidson, Jonathan~Passerat-Palmbach, Giacomo~Tarroni, Amir~Alansary, Ozan~Oktay, Bernhard~Kainz, and~Daniel~Rueckert 
\thanks{All authors are with the Dept. of Computing, Imperial College London, SW7 2AZ, London, UK}
\thanks{(*)Corresponding author: Martin Rajchl e-mail: m.rajchl@imperial.ac.uk}
}


\maketitle

\begin{abstract}
The availability of training data for supervision is a frequently encountered bottleneck of medical image analysis methods. While typically established by a clinical expert rater, the increase in acquired imaging data renders traditional pixel-wise segmentations less feasible. In this paper, we examine the use of a crowdsourcing platform for the distribution of super-pixel weak annotation tasks and collect such annotations from a crowd of non-expert raters. The crowd annotations are subsequently used for training a fully convolutional neural network to address the problem of fetal brain segmentation in T2-weighted MR images. Using this approach we report encouraging results compared to highly targeted, fully supervised methods and potentially address a frequent problem impeding image analysis research. 
\end{abstract}
\begin{IEEEkeywords}
Weak Supervision, Image Segmentation, Machine Learning, Convolutional Neural Networks
\end{IEEEkeywords}

\IEEEpeerreviewmaketitle
\section{Introduction}
\PARstart{M}{odern} learning-based methods for medical image analysis rely on large amounts of labelled data to properly cover different sources of variability in the data (\emph{e.g.} due to the pose of the subject, the presence of pathology, \emph{etc.}). This situation is particularly exacerbated when analysis on data is required for which no open labelled atlas databases exist that could be adopted for supervision. However, the option of an expert rater to pixel-wise label a training set, is often not feasible. To address this problem, methods employing weak forms of annotations (\emph{e.g.} image-level tags \cite{schlegl2015predicting}, bounding boxes \cite{papandreou2015weakly,dai2015boxsup,rajchl2016deepcut}, drawn scribbles \cite{koch2016multi,rajchl2014interactive}, \emph{etc.}) aim to reduce the annotation effort and increasingly gain attention. For instance, recent studies have shown that employing bounding box annotations is approximately 15 times faster than using pixel-wise manual segmentations \cite{lin2014microsoft,papandreou2015weakly}. In conjunction with using simple forms of annotations, web-based collaborative platforms for crowdsourcing have been investigated in their ability to obtain large amounts of annotations for labelling image databases \cite{mckenna2012strategies,maier2014can,haehn2014design}. While such interfaces often have limited capacity to interact with the image data, using weak annotations immediately suggests itself, because of its simplicity. However, in contrast to tasks such as the annotation of natural images \cite{lin2014microsoft,russell2008labelme} and the identification of surgical instruments \cite{maier2014can} in surgical video sequence, the correct interpretation of medical images requires specialised training and experience, and therefore might pose a challenge for non-expert crowds. Nevertheless, in contrast to the \emph{diagnostic interpretation} of medical images, medical image analysis pipelines often require the \emph{identification of anatomical structures}, requiring less expertise (\emph{i.e.} it requires less expertise to identify an organ in an MR image, than a potential pathology). 

\noindent
\subsection{Contributions: }
In this paper, we entertain the notion that non-experts can be used for some annotation tasks on medical images. These tasks can be simplified by employing super-pixel weak annotations and the total annotation effort can be distributed to many raters (also commonly referred to as crowd) using a web browser as an interface. We investigate this concept in the context of the fetal brain segmentation problem in T2-weighted MR images. Using a fully convolutional neural network (FCN) we achieve state-of-the-art accuracy performance under full expert supervision and report comparably high values for learning from expert weakly supervised data (\emph{i.e.} super-pixel annotations). Further, we distribute the super-pixel annotation tasks to 12 non-expert raters and achieve similar performance to that of experts.

\section{Methods}
In the following sections, we describe means of distributing annotation tasks and facilitating learning from acquired weak annotations using a state-of-the-art fully convolutional neural network \cite{long2015fully}. 

\begin{figure*}
\centering
\includegraphics[width=0.95\linewidth]{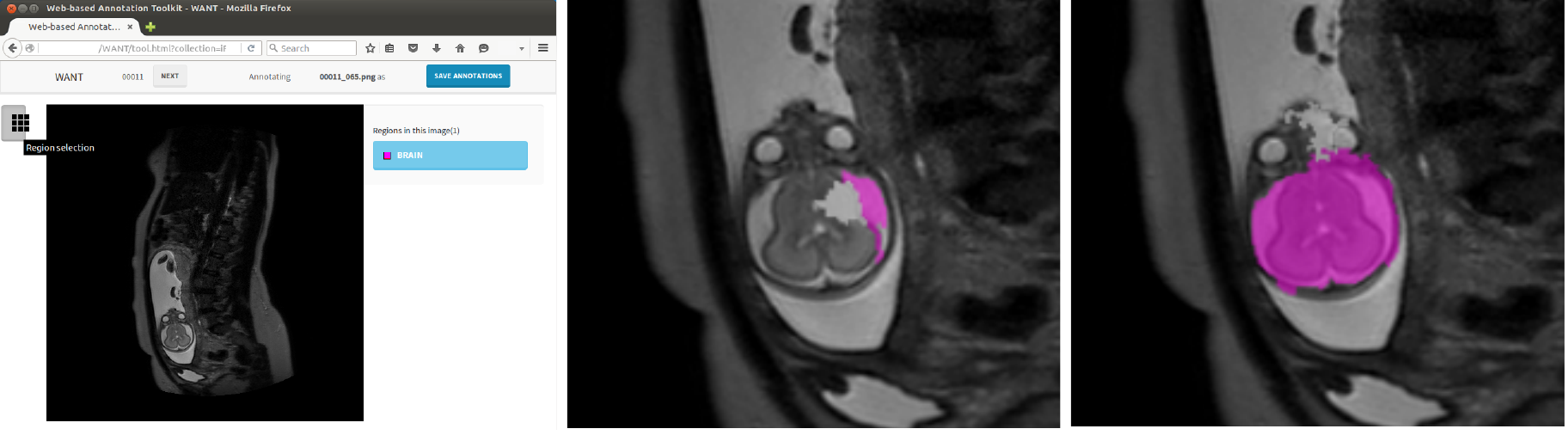}
\caption{Proposed crowdsourcing interface in a browser showing an 
 uterine MR image from a stack of sagittal slices (left). Enlarged examples of progressing super-pixel annotations on the displayed images (middle and right).}
\label{fig:wants}
\end{figure*}

\noindent
\subsection{Distributed Weak Annotations: }
\label{sec:dist_weak_ann}
For a flexible solicitation of annotation tasks, we propose a crowdsourcing platform where users can interact and annotate image data. To accelerate the annotation process, we provide a SLIC super-pixel segmentation \cite{achanta2012slic} and let users select those belonging to the object we are interested in. We implement the SLIC computation using \emph{Javascript} to outsource the computational load to the client machine and concentrate on backend tasks on the server side (\emph{e.g.} data conversion, collection, \emph{etc.}). The web-based user interface is based on the well-known \emph{LabelMe} framework \cite{russell2008labelme} and was modified to interact with volumetric medical image data and to compute and collect super-pixel annotations. Fig. \ref{fig:wants} depicts the interface reduced to accommodate the particular annotation task at hand and an example of a user labelling super-pixel belonging to a fetal brain on a T2w MR image slice.

\noindent
\subsection{Learning with Fully Convolutional Neural Networks: }
\label{sec:learning}
We propose a fully convolutional neural network (FCN) architecture to address the segmentation problem in a general and extendible manner \cite{long2015fully}. Such an approach has recently been introduced for semantic object segmentation problems on natural images, exceeding the state-of-the-art in accuracy performance while exhibiting remarkable training and inference speed thanks to its fully convolutional nature \cite{long2015fully}. 

\begin{figure*}
\centering
\includegraphics[width=0.7\linewidth]{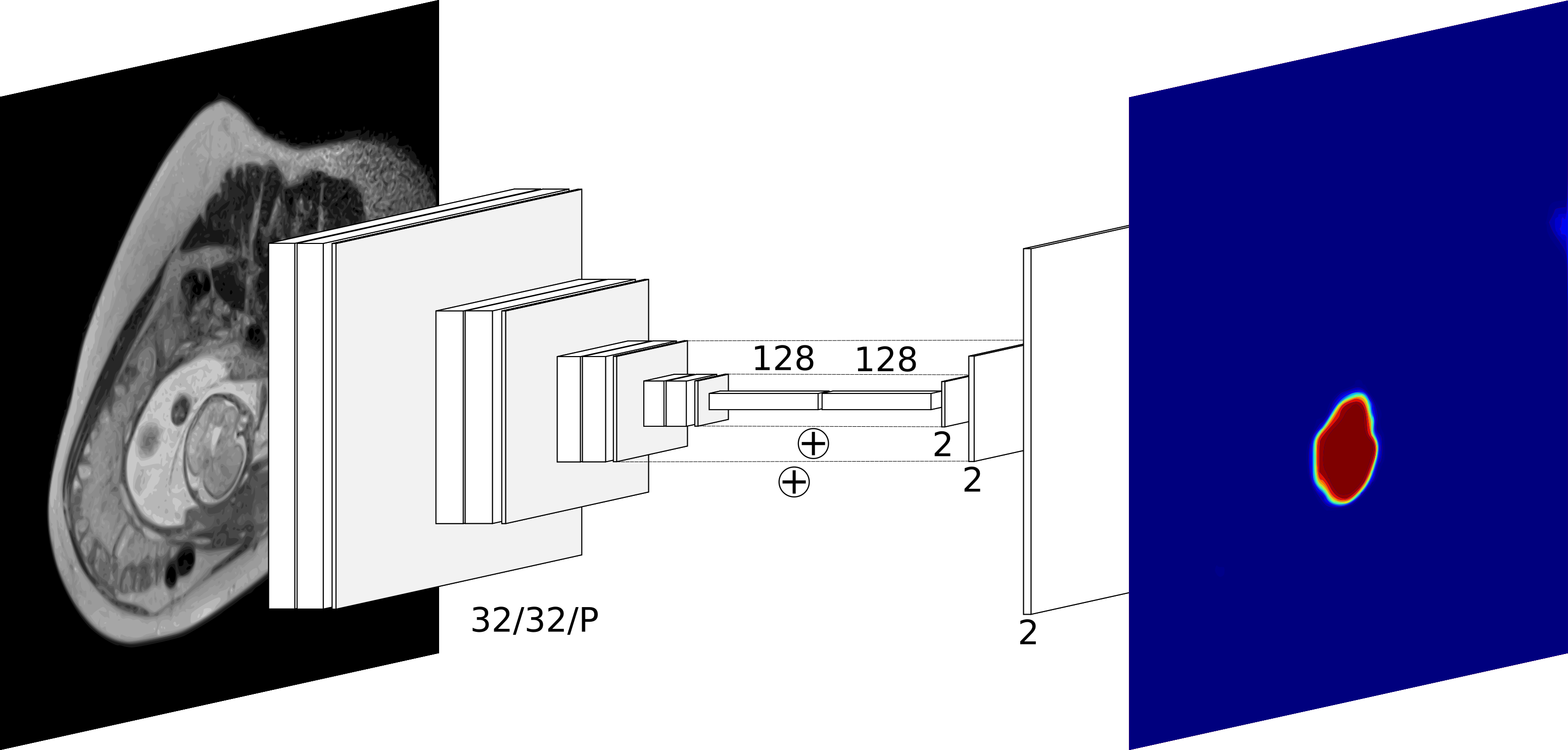}
\caption{Proposed fully convolutional neural network architecture with four stacks of convolutional (white) and max-pooling layers (grey, \emph{P}) and two skip layers to combine outputs from low resolution pooled features with up-sampling deconvolution layers ($\oplus$).}
\label{fig:cnn_arch}
\end{figure*}

\noindent
\subsection{FCN Architecture: }
We designed the network in stacks of two convolutional layers ($L_{Cv}$, $N_{Filters}$ = 32) for feature representation and a max-pooling layer ($L_{MP}$) to downsample the inputs and capture a wider context. In total this combination of layers is repeated four times (using kernel sizes of 5, 5, 5 and 3, respectively) and subsequently connected to two classification layers ($L_{Cl}$, kernel size = 1x1, $N_{Filters}$ = 128). To obtain a prediction of the size of the original input image, we employ backwards strided (de-) convolution layers to upsample the coarse predictions of $L_{Cl}$. Additionally, we employ a skip architecture, where the upsampled coarse semantic prediction scores are combined with appearance information from the feature layers \cite{long2015fully}. In practice this is done with an element-wise summation of the responses of the upsampling layers and corresponding feature layers $L_{MP}$. A positive side-effect of the proposed architecture is that it can take an input of arbitrary size and produce correspondingly-sized output. This is particularly interesting for medical image segmentation problems, where patch-based training is warranted, as object background classes are often highly imbalanced on a pixel level (\emph{e.g.} for fetal brain segmentation, the object occupies only approximately 1.5\% of the image domain).

\noindent
\subsection{FCN Training: }
We train the network via mini-batch stochastic gradient descent on sampled image patches of size $128 \times 128$ with spatially adjacent slices in 3 channels with a fixed learning rate $\alpha$ = 0.01, and a momentum $\mu$ = 0.9 for approximately 20 epochs. In order to prevent parameter over-fitting to the training examples, we incorporate a weight decay of 0.0005 into the gradient computation (to enforce sparsity in the layer responses) and a signal dropout \cite{srivastava2014dropout} of 50\% to all $L_{Cl}$. Before training, all image volumes are normalised to zero mean and unit standard deviation. To offset the class imbalance, we randomly sample an equal amount of patches containing the object and the background (\emph{i.e.} a hybrid over- and undersampling technique). The size of the training sample ($N_{Training}$ = 96000) is set to contain all possible foreground patches in the database and an equal number of background patches. The obtained patches are subjected to data augmentation (to generalise better to unseen data) by incorporating random flips in all directions and by adding a Gaussian distributed intensity offset with the standard deviation $\sigma$ = 1.0 to each patch. This allows to account for residual intensity differences after normalisation between patches of different images. 

\section{Experiments}

\noindent
\subsection{Image Data: }
Images from 37 fetal subjects were acquired on a Philips Achieva 1.5T with the mother lying 20$^{\circ}$ tilt on the left side to avoid pressure on the inferior vena cava or on her back depending on her comfort. Single-shot fast spin echo (ssFSE) T2-weighted sequences are used to acquire stacks of images that are aligned to either the main axes of the fetus or of the mother. Usually three to six stacks are acquired for the whole uterus with a voxel size of 1.25 $\times$ 1.25 $\times$ 2.50 mm. fetal MRI data can be corrupted by motion because of unpredictable fetal movements and maternal respiratory motion. The stacks with the least motion artefacts were selected for our experiments. A clinical expert rater manually annotated the fetal brain to establish a reference standard segmentation. 

\noindent
\subsection{Evaluation: }
We recruited 12 users with technical degrees and exposure to medical imaging research as a non-expert crowd and asked them to annotate consecutive slices of T2w fetal MR volumes using the proposed web interface (see Sec. \ref{sec:dist_weak_ann}) to label super-pixels of size $12 \times 12$ $px$ belonging to a fetal brain. Prior to access to the data, the users were asked to complete a short tutorial showing expert segmentations of the fetal brain in different slice directions. Furthermore, to evaluate the detrimental impact of the SLIC weak annotations, a second experiment was performed using super-pixels extracted from the reference segmentations from the expert rater (based on a threshold of 50\% of area coverage between each super-pixel and the reference). These serve as training data for learning under \emph{expert weak supervision}. Finally, we compare training on \emph{full expert supervision} data (\emph{i.e.} directly from the reference standard) using the proposed FCN architecture in Sec. \ref{sec:learning}. The trained network models are then used to infer the fetal brain on unseen volumes. 
To reduce the variation in the experimental setup and suppress possible factors impacting on accuracy, we sampled the same patch locations for all annotation types and computed the same augmentations. Additionally, prior to training all networks were initialised with the same random weights. For validation, we used a three-fold cross-validation setup and computed the Dice Similarity Coefficient ($DSC = \frac{2 |P \cap M|}{|P|+|M|}$ ) between predicted regions $P$ and expert manual segmentations $M$. We used the Caffe library \cite{jia2014caffe} for the creation and training of the proposed FCN architecture (see Sec. \ref{sec:learning}) and performed all experiments on a Ubuntu 14.04 machine with 256 GB memory and an NVIDIA Tesla K80 (12 GB memory). 

\section{Results}
\begin{figure*}
	\centering
	\includegraphics[width=0.95\linewidth]{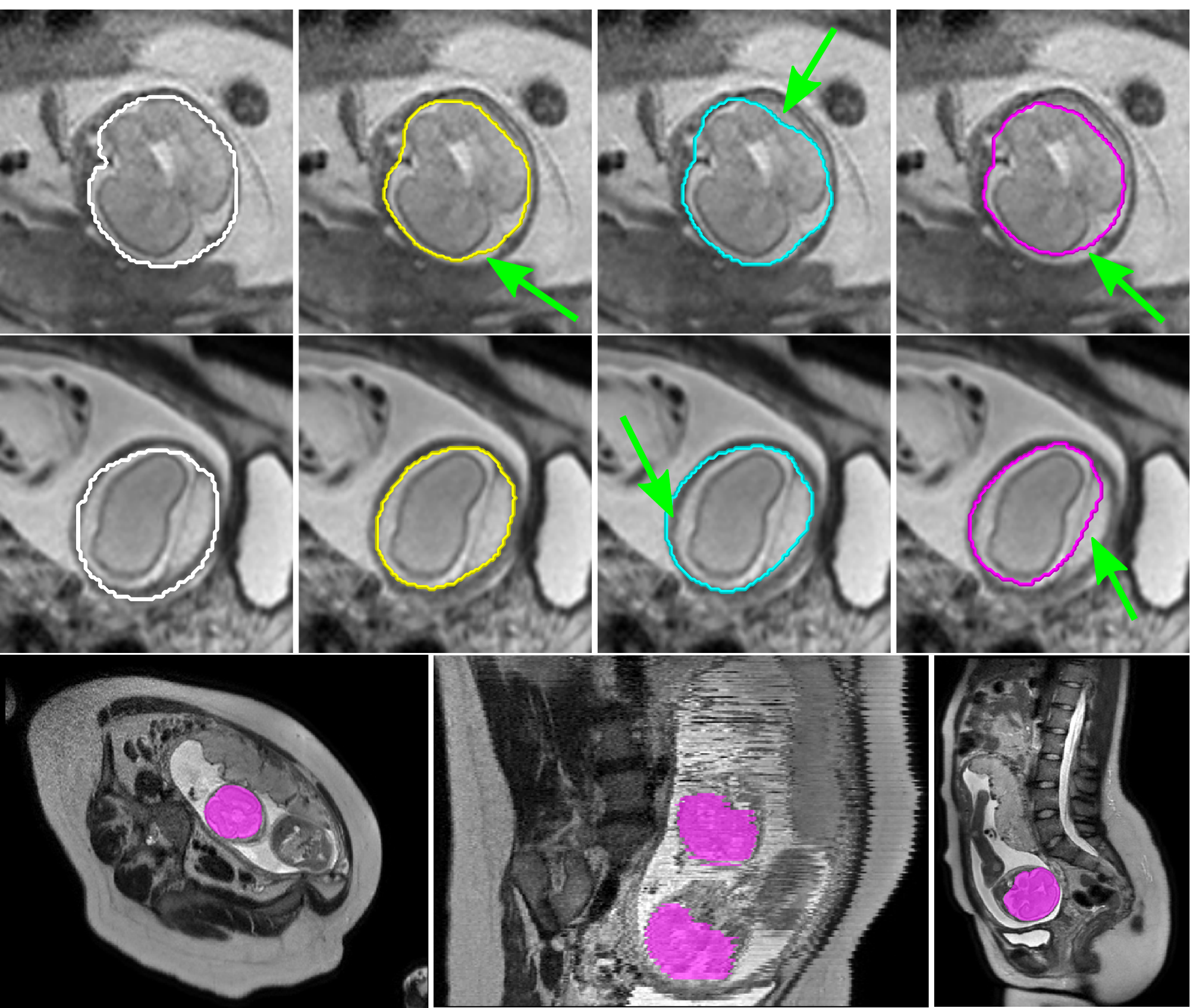}
	\caption{Top and middle row: Example results of expert manual (white), expert fully supervised (yellow), expert weakly supervised (cyan) and non-expert weakly supervised segmentations. Prediction errors are marked with arrows (green). Bottom row: Non-expert weakly supervised segmentation results on axial and sagittal slices, including twins (middle).}
	\label{fig:exampleseg}
\end{figure*}

\noindent
Table~\ref{ta:acc_brain} shows the accuracy as DSC for all compared approaches, respectively. While learning under full expert supervision exhibits the most accurate performance, using weakly supervised simulated expert annotations and weak annotations from non-expert raters yield comparably high results. Figure~\ref{fig:exampleseg} shows selected examples of segmentation results and segmentation errors for all compared methods.

\begin{table}
	\centering
	\caption{\label{ta:acc_brain} Mean accuracy results for fetal brain segmentation compared to the expert reference standard as DSC [\%].}
    \begin{tabular}{p{2cm}ccp{0.5cm}c}
		\toprule
		& \multicolumn{2}{c}{Expert} & & Non-expert  \\
		\midrule
Supervision type & Full & Weak & & Weak \\
		DSC [\%]  & 92.7 $\pm$ 2.3 & 90.3 $\pm$ 2.8 & & 90.6 $\pm$ 2.3\\
		\bottomrule
	\end{tabular}
\end{table}

\begin{figure*}
	\centering
	\includegraphics[width=0.95\linewidth]{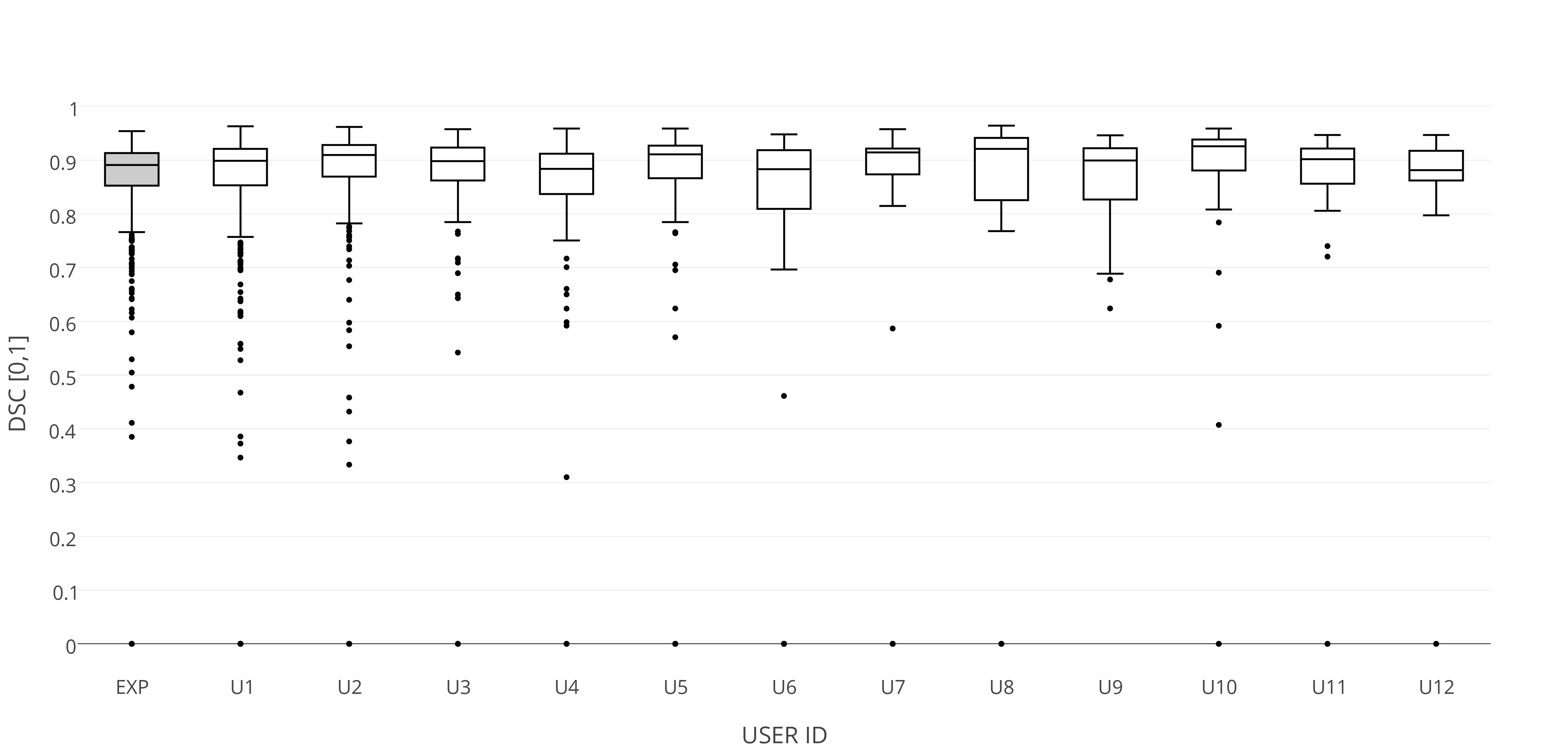}
	\caption{Accuracy of SLIC weak annotations of the simulated expert user (EXP) and non-expert users (U) in DSC [0,1].}
	\label{fig:acc_by_uid}
\end{figure*}

The reliability of the web-based super-pixel annotations collected from non-expert users and those extracted from the expert is depicted in Figure~\ref{fig:acc_by_uid}. All slices that either contained the object in the reference standard or were annotated by a user were evaluated in their accuracy against the reference standard. Note, that the users did \emph{not} perform an equal number of annotation tasks. 

\noindent
\subsection{Runtime: } The average time spent annotating an image slice with the proposed web-interface was 7.2 $\pm$ 3.4 seconds, including loading, annotation and task submission. FCN training time was approximately six hours and inference can be done under one minute for the largest acquired MR stack (512 $\times$ 512 $\times$ 200 vx). Generation of the manual ground truth took approximately three full working days.
		
\section{Discussion}
All compared methods perform well in qualitative comparison with other studies employing highly targeted, fully supervised approaches. The works in \cite{keraudren2014automated}, \cite{Kainz2014sub} and \cite{taleb2013automatic} reported mean DSC scores of 93.0\%, 90.7\% and 80.4\%, respectively. Note, that the FCN prediction could additionally be post-processed with a graphical method, which has been shown to improve results in other segmentation problems \cite{papandreou2015weakly,dai2015boxsup}. As expected, a higher accuracy could be achieved when learning under full supervision, however differences appear marginal compared to those reported using bounding box annotations \cite{papandreou2015weakly,dai2015boxsup}. Surprisingly, both weakly supervised networks present with very similar accuracy (see Tab. \ref{ta:acc_brain}), when random factors such as sampling and augmentation are accounted for. Particularly interesting is the presentation of learned segmentation errors. We expect the exclusion of the cerebro-spinal fluid when using non-expert annotations (\emph{c.f.} Fig. \ref{fig:exampleseg}, magenta) is due to differences in image interpretation of the crowd on where the brain boundary is on axial slices. Similarly, the oversegmentation of the skull (\emph{c.f.} Fig. \ref{fig:exampleseg}, cyan) might be due to systematic oversegmentations from computed expert super-pixels. Systematic annotation errors could be addressed by integration of quality assurance measures and/or annotation regularisation post collection. Considering the base accuracy of the collected fetal brain annotations from non-experts, we observe similar performance to that of an expert (\emph{c.f.} Fig. \ref{fig:acc_by_uid}), indicating that some anatomical annotation tasks can be performed by crowds with less expertise.

The observed efficiency of distributed weak annotation tasks with the proposed crowdsourcing interface is remarkable. Considering the measured average annotation time of 7.2 seconds, with a collective of 12 users, the annotation of the entire database took less than one hour to annotate (total of 10.7 hours) the entire database (more than 5000 slices) an expert annotator took three work days to establish the same with a multi-planar interface. These observations might indicate a paradigm shift on how we enable learning based methods for medical image analysis to address the ever-growing data collected for imaging studies.  

At this juncture, we note that contrary to relying on commercial crowdsourcing platforms such as Amazon MTurk, we aim to focus on a more flexible platform that can better take advantage of contributions from image scientists and those interested in supporting medical research, thereby fostering engagement with a wider general public. The proposed approach has the ability to enable this while simultaneously enabling the development of machine learning based methods at much larger scales.

\subsection{Conclusions: }
We have investigated the web-based distribution of weak annotation tasks to a crowd of non-expert users to establish training data for a learning-based segmentation method. The proposed approach largely reduces the annotation load on expert users and was successfully employed for segmentation of fetal brains from motion-corrupted T2w MR image stacks. The encouraging results and the consistent annotation performance of the crowd suggest that this approach could be readily ported to other challenges and potentially address a frequently encountered bottleneck in medical image analysis studies.

\section*{Acknowledgements}
We gratefully acknowledge the support of NVIDIA Corporation with the donation of a Tesla K40 GPU used for this research. This research was also supported by the National Institute for Health Research (NIHR) Biomedical Research Centre based at Guy's and St Thomas' NHS Foundation Trust and King's College London.  The views expressed are those of the author(s) and not necessarily those of the NHS, the NIHR or the Department of Health.  Furthermore, this work was supported by Wellcome Trust and EPSRC IEH award [102431] for the iFIND project and the Developing Human Connectome Project, which  is funded through a Synergy Grant by the European Research Council (ERC) under the European Union's Seventh Framework Programme (FP/2007-2013) / ERC Grant Agreement number 319456.

\appendices

\bibliographystyle{IEEEtran}
\bibliography{refs}

\end{document}